\def\year{2020}\relax
\documentclass[letterpaper]{article} 
\usepackage{aaai20}  
\usepackage{times}  
\usepackage{helvet} 
\usepackage{courier}  
\usepackage[hyphens]{url}  
\usepackage{graphicx} 
\usepackage{graphicx}  
\usepackage{tikz}
\usepackage{booktabs}
\usepackage{amssymb}
\usepackage{multirow}
\usepackage{enumerate}
\definecolor{mygreen}{cmyk}{1,0,1,0.25}
\definecolor{myblue}{cmyk}{1,1,0,0}
\definecolor{myred}{cmyk}{0,1,1,0}
\newcommand{\citet}[1]{\citeauthor{#1} (\citeyear{#1})}
\title{An Annotated Corpus of Reference Resolution \\for Interpreting Common Grounding}
\author{Takuma Udagawa\\
{The University of Tokyo, Tokyo, Japan}\\
{takuma\_udagawa@nii.ac.jp}
\And
Akiko Aizawa\\
{National Institute of Informatics, Tokyo, Japan}\\
{aizawa@nii.ac.jp}
}
 
\begin{document}

\maketitle

\begin{abstract}
Common grounding is the process of creating, repairing and updating mutual understandings, which is a fundamental aspect of natural language conversation. However, interpreting the process of common grounding is a challenging task, especially under continuous and partially-observable context where complex ambiguity, uncertainty, partial understandings and misunderstandings are introduced. Interpretation becomes even more challenging when we deal with dialogue systems which still have limited capability of natural language understanding and generation. To address this problem, we consider reference resolution as the central subtask of common grounding and propose a new resource to study its intermediate process. Based on a simple and general annotation schema, we collected a total of 40,172 referring expressions in 5,191 dialogues curated from an existing corpus, along with multiple judgements of referent interpretations. We show that our annotation is highly reliable, captures the complexity of common grounding through a natural degree of reasonable disagreements, and allows for more detailed and quantitative analyses of common grounding strategies. Finally, we demonstrate the advantages of our annotation for interpreting, analyzing and improving common grounding in baseline dialogue systems.
\end{abstract}

\section{Introduction}
\label{section:introduction}

\begin{figure*}[tb!]
\centering
\includegraphics[width=0.96\textwidth]{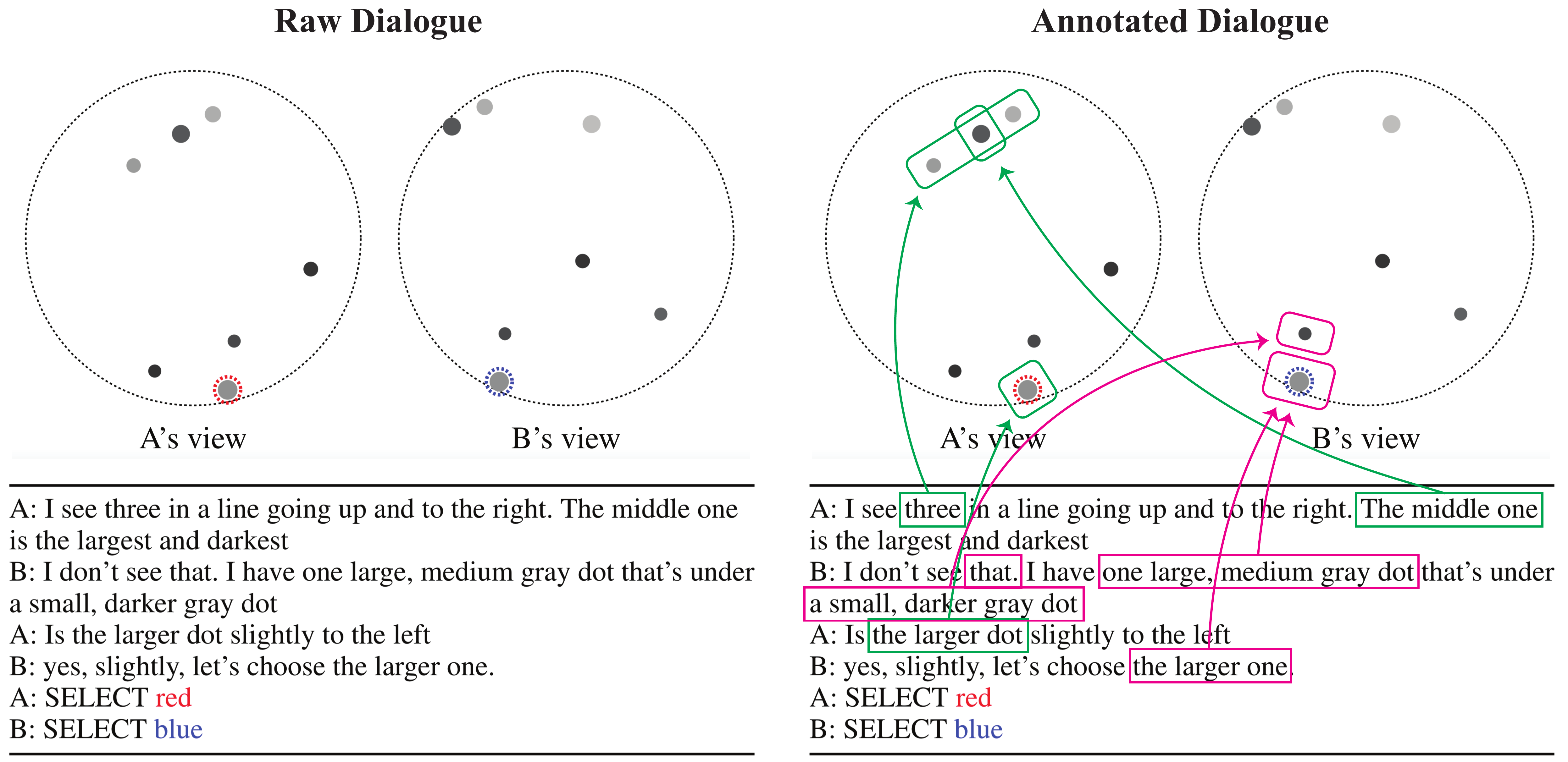}
\caption{A visualized example of the raw dialogue (left) and our annotated dialogue (right). In our annotation, referring expressions are detected and their intended referents are annotated based on the speaker's view (only one judgement shown in this example). Background task is described in detail in Section \ref{section:background_task} and our annotation procedure in Section \ref{section:annotation_procedure}.
}
\label{fig:first_example}
\end{figure*}

Common grounding is the process of creating, repairing and updating mutual understandings, which is a critical aspect of sophisticated human communication \cite{clark1996using} as well as a longstanding goal in dialogue modeling \cite{traum1994computational}. Recently, there have been several new proposals of dialogue tasks which require advanced skills of common grounding under \textit{continuous} and \textit{partially-observable} context \cite{udagawa2019natural,haber-etal-2019-photobook}. Their main contributions include proposal of clear evaluation metrics based on task success rate, collection of large-scale datasets (thousands of dialogues) and introduction of complex ambiguity, uncertainty, partial understandings and misunderstandings which are minimally observed under traditional settings based on either categorical or fully-observable context.

However, interpretation of the process of common grounding remains largely an open problem. Although a formal theory such as \citet{poesio2010completions} can account for some of the important details in common grounding, constructing such precise semantic representation is a difficult and costly process, especially under continuous and partially-observable context with high ambiguity and uncertainty. Interpretation becomes even more challenging when we deal with dialogue systems represented by end-to-end neural models \cite{vinyals2015neural,bordes2016learning}, which can converse fluently but still lack true competency of natural language understanding and generation.

In this work, we approach this problem by \textit{decomposing} the common grounding task based on its intermediate subtasks. Specifically, we consider \textit{reference resolution} as the central subtask of common grounding (in the sense that mutual understanding can only be created through successful references to the entities in the task domain), define this subtask formally based on a simple and general annotation schema, and create a large-scale resource to study this subtask along with the original task of common grounding.

Our annotated corpus consists of a total of 40,172 referring expressions in 5,191 dialogues curated from the existing corpus \cite{udagawa2019natural}, along with multiple (a minimum of 3) judgements for referent interpretations. A visualization of our annotation is shown in Figure \ref{fig:first_example}.

Through our corpus analysis, we show that our annotation has high agreement in general but also includes a natural degree of reasonable disagreements, which verified that our annotation can be conducted reliably while capturing the ambiguity and uncertainty under continuous and partially-observable context. In addition, we give a more quantitative analysis of \textit{pragmatic expressions} as an illustrative example of analyses that can be conducted based on our annotation.

Finally, through our experiments we show that our annotation is critical for interpreting and analyzing common grounding in baseline dialogue systems, as well as improving their performance on difficult end tasks.

Overall, we propose a fundamental method and resource for interpreting the process of common grounding through its subtask of reference resolution. All materials related to this work will be publicly available at \url{https://github.com/Alab-NII/onecommon}.

\section{Related Work}
\label{section:related_work}

One of the most influential models of common grounding to date is the \textit{contribution} model \cite{clark1996using}, which distinguishes information in a dialogue into two phases: the \textit{presentation phase} where a piece of information is introduced by a speaker, and the \textit{acceptance phase} where it gets accepted by a listener. However, applying such theory for analysis in realistic settings can be difficult or even problematic \cite{koschmann2003reconsidering}, especially when contributions are implicit, indirect, unstructured, uncertain or partial. In contrast, we propose a more practical approach of decomposing common grounding based on well-defined subtasks: in our case we focus on reference resolution. Although our approach does not give a formal account of common grounding, we show that our annotation is \textit{general} with simple and clear definition, \textit{reliable} in terms of annotator agreement under complex settings, and \textit{useful} for interpreting and analyzing the intermediate process of common grounding.

Our work is also relevant to the recent literature of interpretable and explainable machine learning \cite{doshi2017towards,lipton2016mythos}. Especially the analysis of neural based models is gaining attention in NLP \cite{belinkov-glass-2019-analysis}, including end-to-end dialogue models \cite{sankar-etal-2019-neural}. The main novelty of our approach is that we decompose the original task (\textit{common grounding}) based on its central subtask (or could be subtasks), define the subtask (\textit{reference resolution}) formally with an annotation framework, and create a large-scale resource to study the subtask along with the original task. Our approach has several advantages compared to previous analysis methods. First, it is applicable to \textit{both humans and machines}, which is especially important in dialogue domains where they interact. Second, it can be used to study the \textit{relationships} between the original task and its subtasks, which is critical for a more \textit{skill-oriented} evaluation of artificial intelligence \cite{Hernndez-Orallo:2017:MME:3110808,sugawara2017prerequisite}. Third, it can be used for investigating \textit{the dataset} on which the models are trained: this is important in many aspects, such as understanding undesirable bias in the dataset \cite{gururangan-etal-2018-annotation,sugawara-etal-2018-makes} or correct model predictions based on the \textit{wrong reasons} \cite{mccoy-etal-2019-right}. Finally, the collected resource can be used for both \textit{probing} whether the models solve the subtasks implicitly \cite{linzen2016assessing} or \textit{developing} new models which can be explicitly supervised, evaluated and interpreted based on the subtasks.

Coreference and anaphora resolution have also been studied extensively in NLP \cite{pradhan2011conll,poesio2016anaphora}, including disagreements in their interpretations \cite{recasens-etal-2012-annotating,poesio-etal-2019-crowdsourced}. The main difference between our annotation schema and theirs is that we focus on exophoric references and directly annotate the referent entities of each referring expression in situated dialogues. We show that our annotation can be conducted reliably, even by using non-expert annotators for referent identifications. Our annotation does not capture explicit relations between anaphora, but they capture basic coreference relations as well as complex associative anaphora (such as \textit{part-of} relations), at least in an indirect way. Most importantly, they are compatible with such existing schema, and annotating explicit anaphoric relations could be a viable approach for future work.

Finally, visually grounded dialogues have been studied in a wide variety of settings. In comparison, the main strengths and novelty of our corpus can be summarized as follows:

\begin{enumerate}[A.]
  \item Our corpus is based on the advanced setting of continuous and partially-observable context where complex common grounding strategies are required.
  \item Our corpus has more simplicity and controllability compared to realistic visual dialogues, which makes controlled experiments and analyses easier.
  \item Our corpus includes large-scale manual annotation of reference resolution and detailed analyses of agreements/disagreements based on multiple judgements.
\end{enumerate}

Prior work in common grounding \cite{potts2012goal,de2017guesswhat} and visual reference resolution \cite{tokunaga-etal-2012-rex,zarriess-etal-2016-pentoref,shore-etal-2018-kth} mostly focus on categorical or fully-observable settings and do not satisfy A. While visual dialogues \cite{das2017visual,haber-etal-2019-photobook,chen2019touchdown,ilinykh2019meetup} have the strengths of being more complex and realistic, they do not satisfy B and C. Although \citet{gotze-boye-2016-spaceref} conducted a smaller-scale (and more loosely defined) annotation of reference resolution, they did not assess the reliability of the annotation (hence does not satisfy B and C). To the best of our knowledge, our work is the first to satisfy all of the above criteria.

\section{Background Task}
\label{section:background_task}

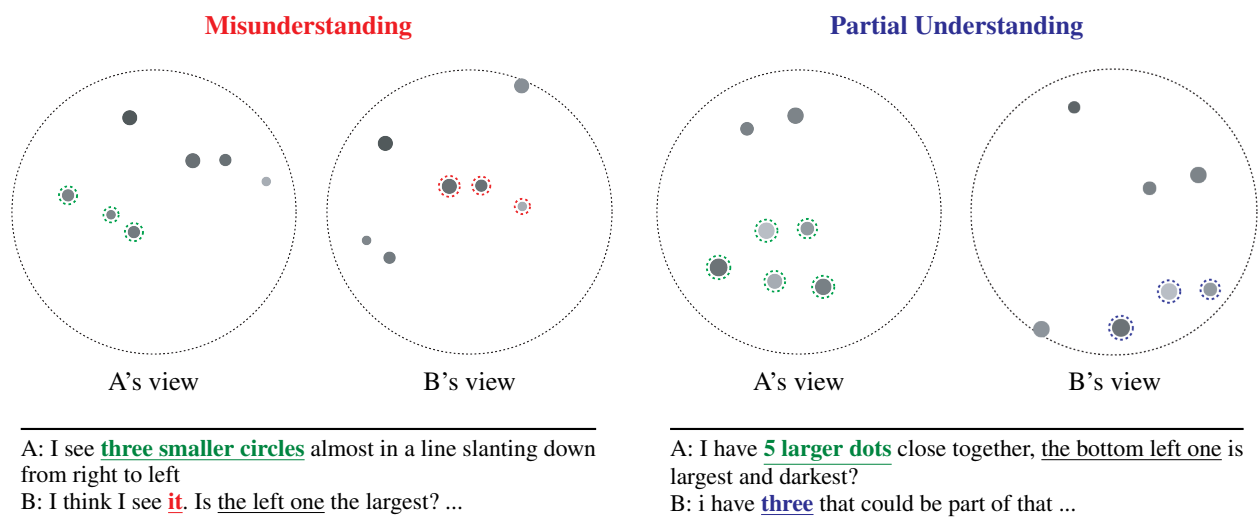
\begin{figure*}[tb!]
\centering
\begin{tabular}{cc}
	{\color{myred} \textbf{Misunderstanding}} & {\color{myblue} \textbf{Partial Understanding}} \\
	\rule{0pt}{30ex}
	\begin{tikzpicture}
	\node[inner sep=0pt] (agent_0) at (0,0)
	  {\includegraphics[width=0.47\columnwidth]{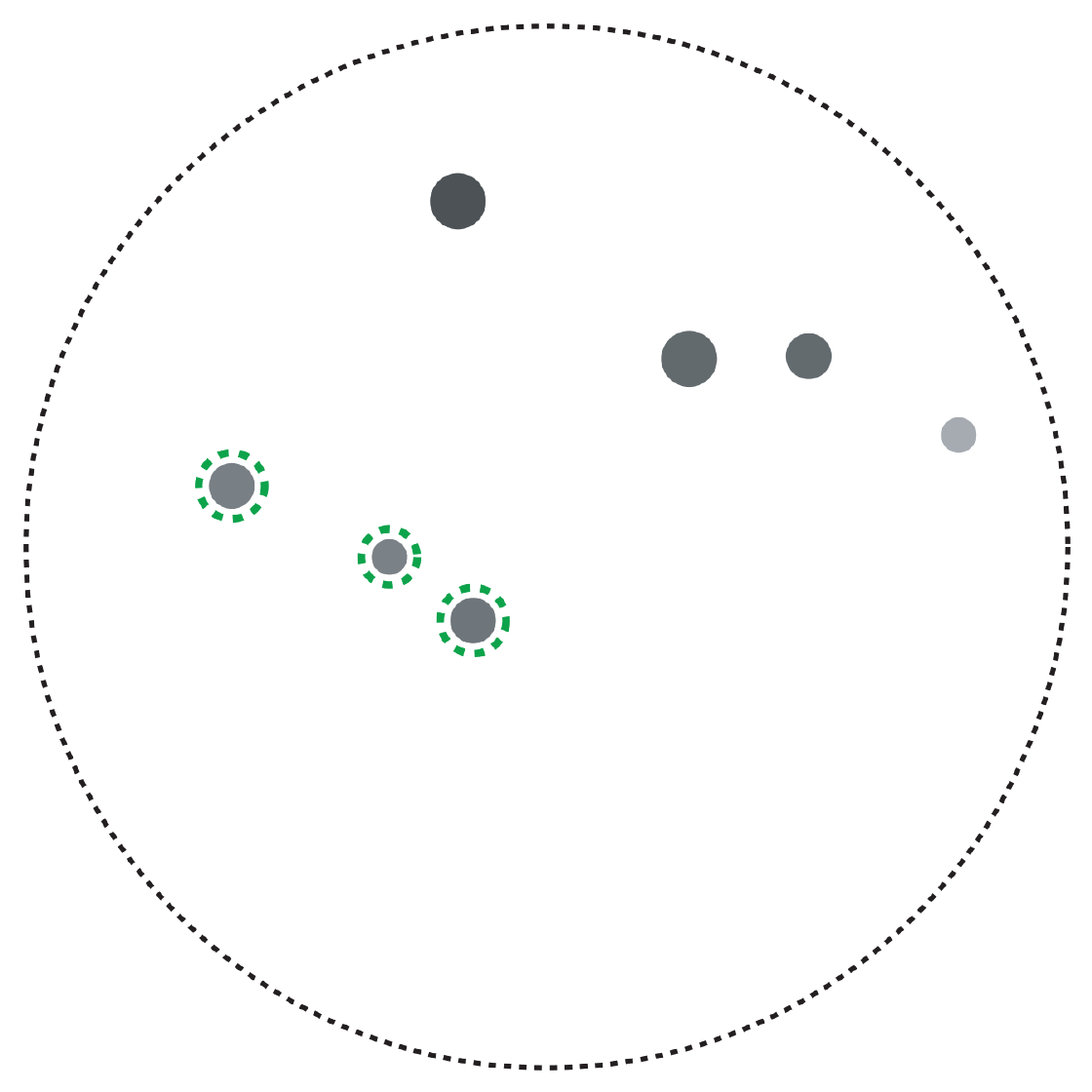}};
	\node[inner sep=0pt] (agent_1) at (4.2,0)
	  {\includegraphics[width=0.47\columnwidth]{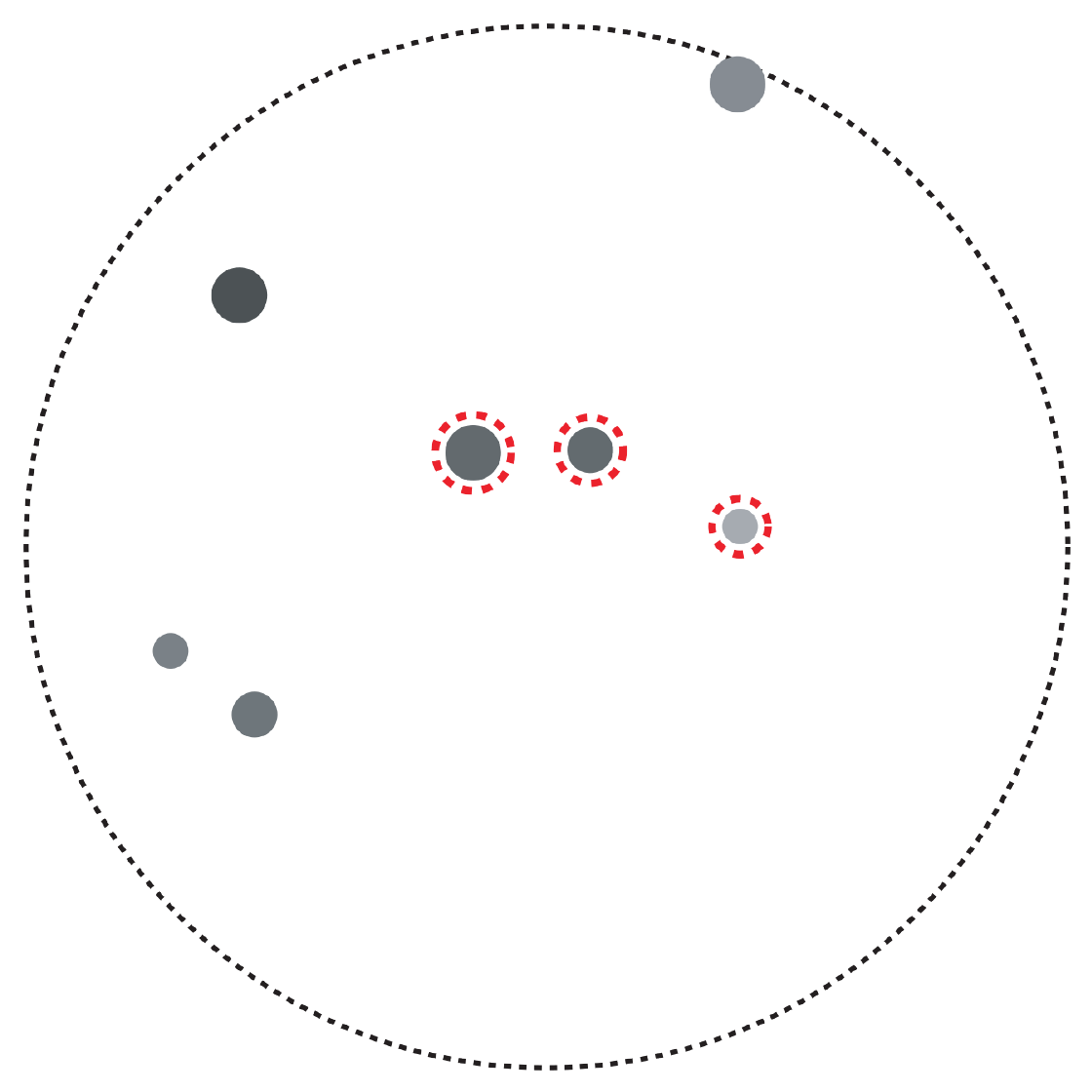}};
	\node [below] at (0,-2) {A's view};
	\node [below] at (4.2,-2) {B's view};
	\end{tikzpicture}
	&
	\begin{tikzpicture}
	\node[inner sep=0pt] (agent_0) at (0,0)
	  {\includegraphics[width=0.47\columnwidth]{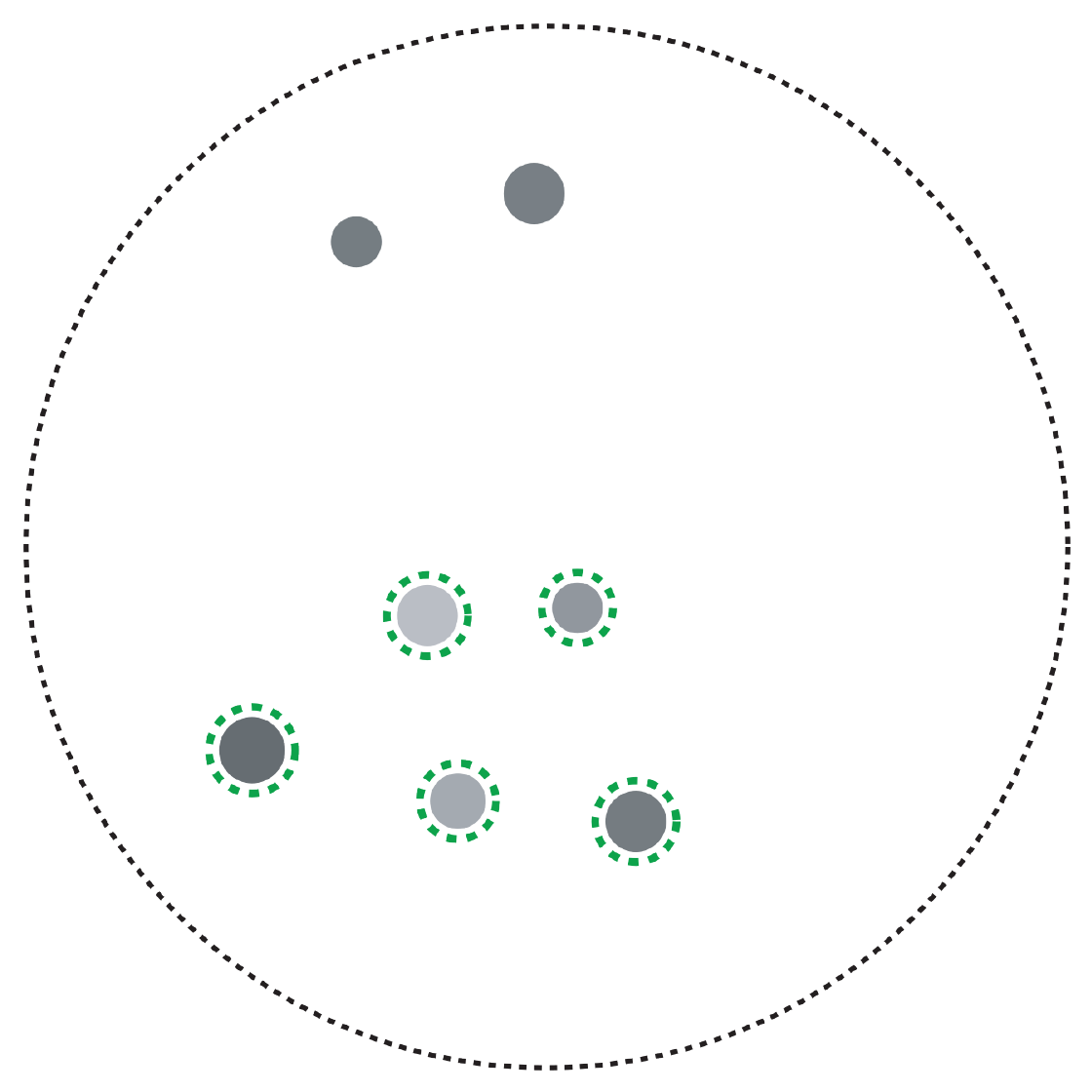}};
	\node[inner sep=0pt] (agent_1) at (4.2,0)
	  {\includegraphics[width=0.47\columnwidth]{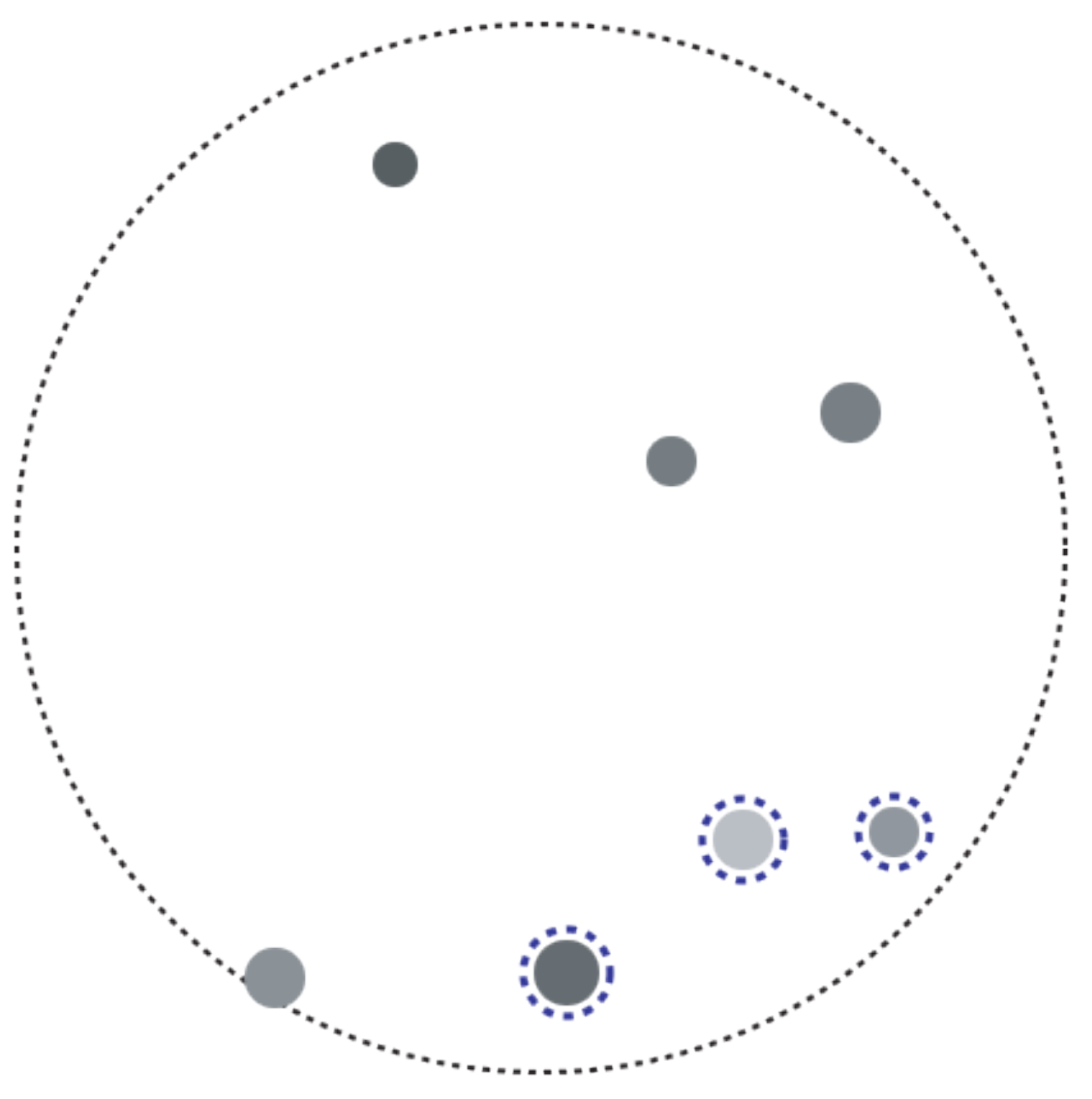}};
	\node [below] at (0,-2) {A's view};
	\node [below] at (4.2,-2) {B's view};
	\end{tikzpicture} \\
	\small
	\begin{tabular}[t]{@{}l@{}}
	\toprule
	A: I see {\color{mygreen} \underline{\textbf{three smaller circles}}} almost in a line slanting down\\
	from right to left \\
	B: I think I see {\color{myred} \underline{\textbf{it}}}.  Is \underline{the left one} the largest? ... \\
	\bottomrule
	\end{tabular}
	&
	\small
	\begin{tabular}[t]{@{}l@{}}
	\toprule
	A: I have  {\color{mygreen} \underline{\textbf{5 larger dots}}} close together, \underline{the bottom left one} is \\
	largest and darkest? \\
	B: i have {\color{myblue} \underline{\textbf{three}}} that could be part of that ... \\
	\bottomrule
	\end{tabular}
\end{tabular}
\caption{Example of misunderstanding and partial understanding captured by our annotation.
}
\label{fig:misunderstanding_partial_understanding}
\end{figure*}

Our annotation is conducted on a recently proposed common grounding dataset, which is a minimal formalization of a collaborative referring task under \textit{continuous} and \textit{partially-observable} context \cite{udagawa2019natural}. In this task, two players are given slightly different but overlapping perspectives of a 2-dimensional grid. Both players have 7 entities in each view, but only 4, 5 or 6 of them are in common: this makes their setting \textit{partially-observable} with different degrees of partial-observability. In addition, each entity only has \textit{continuous} attributes (color, size and location). The goal of the task is to find one of the common entities through natural language communication, and the task is successful if and only if they could find and select the same entity.

Some distinguishing characteristics of their dataset include its large size (a total of 6,760 dialogues, out of which 5,191 were successful on the task), rich linguistic variety with limited vocabulary (a total of only 2,035 unique tokens after preprocessing in our curated corpus), and most importantly the complexity of common grounding introduced by continuous and partially-observable context. As shown in Figure \ref{fig:misunderstanding_partial_understanding}, there could be complex misunderstandings and partial understandings that need to be resolved through advanced skills of common grounding. We can also find various nuanced expressions (\textit{``almost in a line"}, \textit{``I think I see ..."}, \textit{``could be"}) and pragmatic expressions (\textit{``a line"}, \textit{``largest"}, \textit{``bottom left"}) which can be ambiguous or need to be interpreted based on their context.

\section{Annotation Procedure}
\label{section:annotation_procedure}

The goal of our annotation is to provide a \textit{general}, \textit{reliable} and \textit{useful} annotation of reference resolution to interpret the intermediate process of common grounding. In this work, we use the 5,191 successful dialogues from the existing corpus which are expected to be of higher quality (however, our annotation is applicable to unsuccessful dialogues as well). Our annotation procedure consists of two main steps: \textit{markable detection} to semi-automatically annotate referring expressions currently under consideration and \textit{referent identification} to identify the referents of each referring expression.

As an optional step, we also conducted \textit{preprocessing} of the dialogues to correct obvious misspellings and grammatical errors. Due to the limited size of the vocabulary, we manually looked for rare unigrams and bigrams in the dialogue and carefully created rules to correct them. Our preprocessing step is reversible, so the collected annotation can also be applied to the original dialogues without preprocessing.

\subsection{Markable Detection}
\label{subsection:markable_detection}

In this work, we define a \textit{markable} as an independent referring expression of the entities currently under consideration (in our case, the dots in the circular view). Basically, we annotate a markable as a minimal noun phrase including all prenominal modifiers (such as determiners, quantifiers, and adjectives) but excluding all postnominal modifiers (such as prepositional phrases and relational clauses). This eliminates the complexity of the annotation because markables will not overlap or nest with each other. See the figures for many examples of the detected markables (underlined).

To reduce the annotation effort in the later process, we optionally annotate three attributes for each markable if they are obvious from the context: a \textit{generic} attribute when the markable is not specific enough to identify the referents, \textit{all-referents} when the markable is referring to all of the entities in the speaker's view, and \textit{no-referent} when the referents are empty. \textit{Generic} markables are ignored in our annotation, and the referents of \textit{all-referents} or \textit{no-referent} are annotated automatically in the later process. To reduce the redundancy of annotation, we consider a predicative noun phrase as a markable only if there is no previous markable in the same utterance that refer to the same entities: for example, \textit{``a triangle"} in \textit{``\underline{three dots} are forming a triangle"} is not considered as a markable since \textit{``three dots"} is already annotated, but it is considered a markable in \textit{``\underline{one light dot} and \underline{two dark dots} are forming \underline{a triangle}"}. We also annotate obvious \textit{anaphoric} and \textit{cataphoric} relations in the same utterance: this way, the referents of anaphoric and cataphoric markables can be annotated automatically based on their antecedents or postcedents. Note that we do not annotate such relations across utterances, as they can actually refer to different entities (see Figure \ref{fig:misunderstanding_partial_understanding} for such example).

Detection of the markables, their attributes and relations are conducted using the brat annotation tool \cite{stenetorp2012brat}. Annotators were trained extensively and had access to all available information (including original dialogues, players' observations and selections) during annotation.

\subsection{Referent Identification}
\label{subsection:referent_identification}

\begin{figure*}[tb!]
\centering
\includegraphics[width=0.86\textwidth]{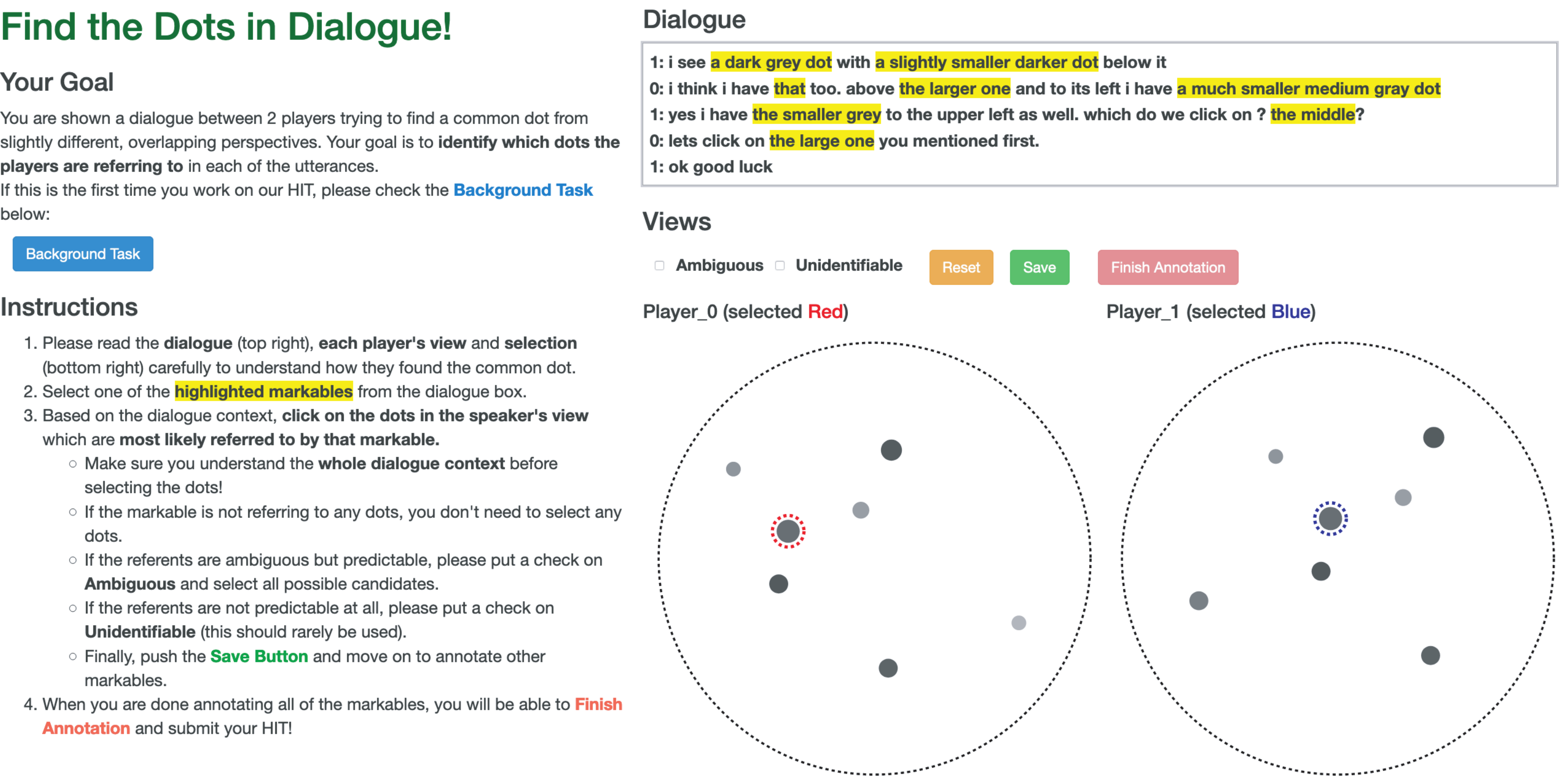}
\caption{Visual interface for referent identification.}
\label{fig:referent_identification_interface}
\end{figure*}

Next, we used crowdsourcing on Amazon Mechanical Turk to collect large-scale judgements of the referents of each markable. Our visual interface for referent identification is shown in Figure \ref{fig:referent_identification_interface}. Annotators were instructed to read the instructions carefully (including description of the background task), put a check on \textit{ambiguous} box and select all possible candidates when the referents are ambiguous, and put a check on \textit{unidentifiable} if the referents are completely unidentifiable based on the available information.

To collect reliable annotations, we restricted the workers to those with at least 100 previously completed HITs and above 99\% acceptance rate. We paid the workers well, with \$0.25 for dialogues with less than 7 markables, \$0.35 with less than 14 markables, and \$0.45 otherwise. In addition, we automatically detected outliers based on several statistics (such as agreement with other workers) and manually reviewed them to encourage better work or reject clearly unacceptable works. The overall rejection rate was 1.18\%.

As a result of this careful crowdsourcing, we were able to collect a large-scale annotation of 103,894 judgements with at least 3 judgements for each of the 34,341 markables.

\section{Annotated Corpus}
\label{section:annotated_corpus}

\subsection{Basic Statistics}
\label{subsection:agreement_statistics}

\begin{table*}[htb!]
\centering \small
\begin{tabular}{ccccccc}
\toprule
\# Markables & \# All-Referents & \# No-Referent & \# Anaphora & \# Cataphora & \% Start Agreement & \% End Agreement \\
\midrule
40,172 & 128 & 1,149 & 4,548 & 6 & 99.11 (96.32) & 99.06 (96.11) \\
\bottomrule
\end{tabular}
\caption{\label{markable_detection_statsitics}
Basic statistics of markable detection. Referents for \textit{all-referents}, \textit{no-referent}, \textit{anaphora} and \textit{cataphora} are annotated automatically. 130 dialogues with 3 independent annotations are used to compute agreement (Fleiss's Multi-$\pi$ in parenthesis).
}
\end{table*}

\begin{table*}[htb!]
\centering \small
\begin{tabular}{cccccc}
\toprule
\# Markables & \# Judgements & \% Ambiguous & \% Unidentifiable & \% Agreement & \% Exact Match\\
\midrule
34,341 & 103,894 & 4.65 & 0.77 & 96.26 (88.66) & 86.90 \\
\bottomrule
\end{tabular}
\caption{\label{referent_identification_statistics}
Basic statistics of referent identification, along with the rate of \textit{ambiguous} and \textit{unidentifiable} checked in the judgements. Agreement is calculated at the entity level (Fleiss's Multi-$\pi$ in parenthesis) and exact match rate at the markable level.
}
\end{table*}

First, we report the basic statistics of the annotation for markable detection in Table \ref{markable_detection_statsitics} and referent identification in Table \ref{referent_identification_statistics}. All agreements are computed based on pairwise judgements. For markable detection, agreement is calculated for the markable text span (at the token level of whether each token is the start or end of the markables). Agreements for markable attributes and relations are also publicly available (but omitted in this paper since they were optional and annotated only in obvious cases). For referent identification, agreement is calculated based on binary judgements of whether each entity is included in the referents or not, and exact match is calculated only if the referents of the markable matched exactly. In addition, we compute Fleiss's Multi-$\pi$ \cite{fleiss1971measuring} to remove the effect of chance level agreements.

Overall, we found high agreement for all annotations, which verified the reliability of our annotation framework.

\subsection{Disagreement Analysis}
\label{subsection:disagreement_analysis}

\begin{figure}[htb!]
\begin{tikzpicture}
\node[inner sep=0pt] (agent_0) at (0,0)
  {\includegraphics[width=0.23\textwidth]{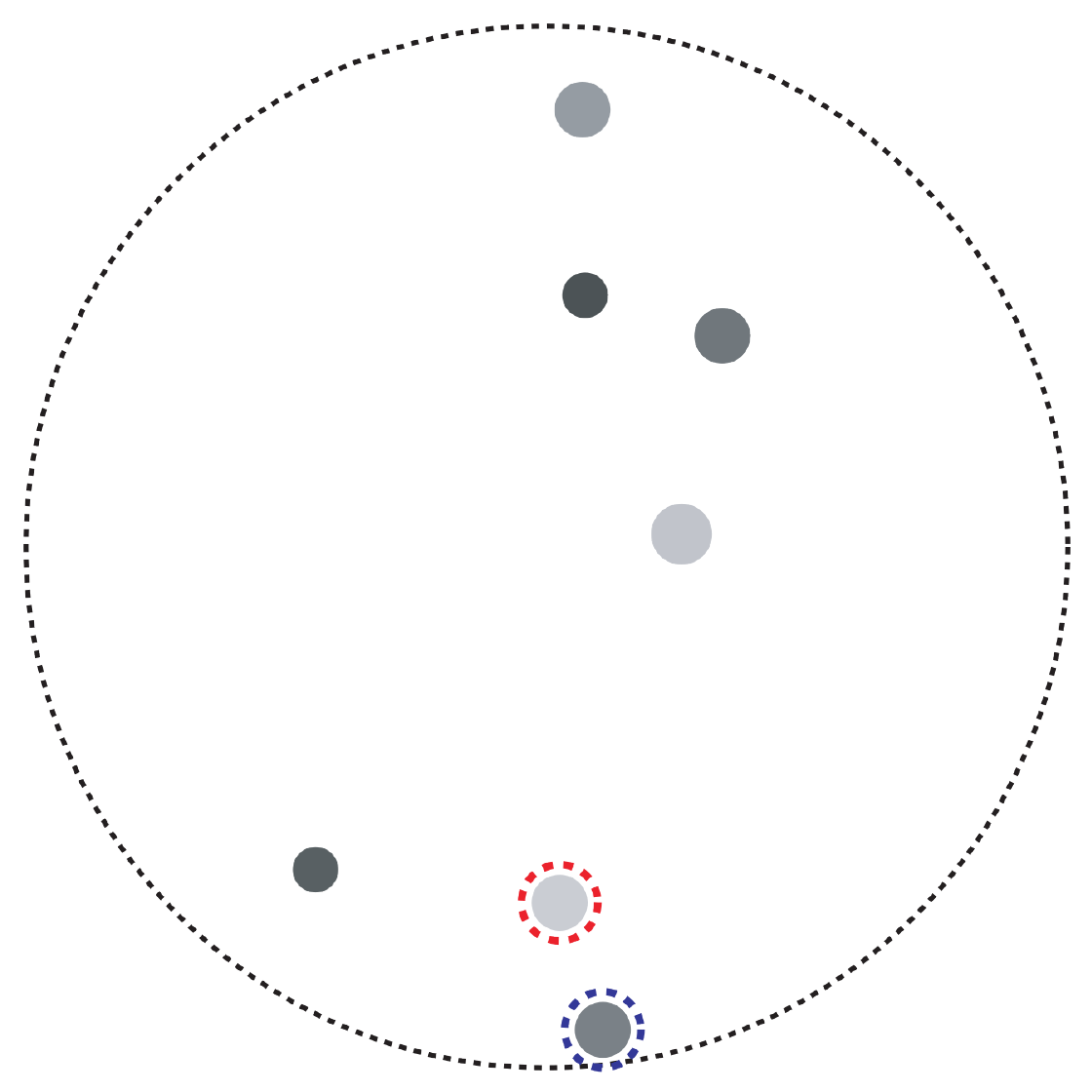}};
\node[inner sep=0pt] (agent_1) at (4.2,0)
  {\includegraphics[width=0.23\textwidth]{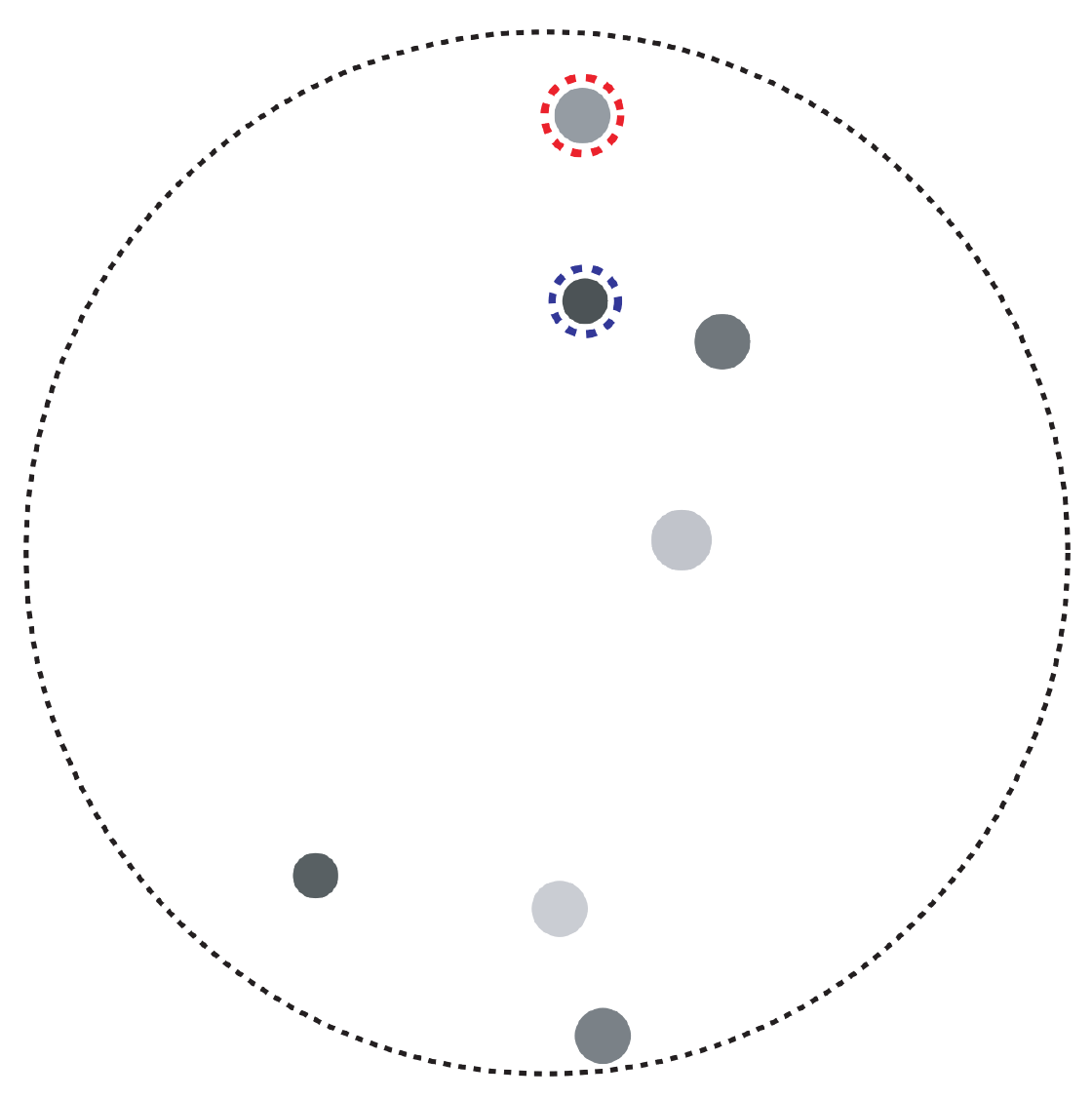}};
\node [below] at (0,2.6) {Annotator 1};
\node [below] at (4.2,2.6) {Annotator 2};
\node[text width=8cm] [below] at (2.1,-2.2) {{\color{myred} \underline{\textbf{medium sized light gray dot}}} with {\color{myblue} \underline{\textbf{a darker one}}} directly under {\color{myred} \underline{\textbf{it}}} and to the right? \\};
\end{tikzpicture}
\caption{Example of seemingly reasonable disagreements captured by our annotation.
}
\label{fig:reasonable_disagreement}
\end{figure}

However, it is natural that there is a certain degree of disagreements in referent interpretations. In fact, it is important to capture such disagreements as there can be genuine ambiguity and uncertainty under continuous and partially-observable context (see Figure \ref{fig:reasonable_disagreement} for example). Therefore, in addition to \textit{explicitly} annotating the ambiguity and unidentifiability as described in Subsection \ref{subsection:referent_identification}, we aim to capture them \textit{implicitly} by collecting multiple judgements from different annotators, similar in approach to \citet{poesio-etal-2019-crowdsourced}.

To study the disagreements in detail, we compute the observed agreement statistics given \textit{the number of referents} in each judgement. To be specific, for a certain number of referents (from 0 to 7), we consider all judgements with the number of referents, make all possible pairs with other judgements on the same markable, and compute the average of entity level agreement and exact match rate. The results are summarized in Table \ref{num_referent_agreement}.

\begin{table}[htb!]
\centering \small
\begin{tabular}{c|ccc}
\toprule
\# Referents & \% Agreement & \% Exact & \% Judgements \\ 
\midrule
0 & 78.04 & 17.78 & \phantom{0}1.31 \\  
1 & 97.45 & 90.28 & 71.81 \\  
2 & 94.87 & 82.17 & 14.85 \\  
3 & 93.93 & 83.03 & \phantom{0}7.51 \\  
4 & 92.18 & 76.66 & \phantom{0}2.20 \\  
5 & 90.31 & 71.03 & \phantom{0}0.88 \\  
6 & 90.75 & 78.14 & \phantom{0}1.22 \\  
7 & 81.47 & 62.50 & \phantom{0}0.21 \\  
\bottomrule
\end{tabular}
\caption{\label{num_referent_agreement}
Agreement statistics given the number of referents in the judgement and the percentages of such judgements.
}
\end{table}

We can see that there is a significant amount of disagreements when the number of referents was judged to be either 0 or 7. This could be due to several reasons: obvious cases were already annotated as \textit{no-referent} or \textit{all-referents} during markable detection (so only difficult cases were left), annotators simply made mistakes (e.g. forgot to annotate), or the referents were annotated as such when it was too difficult to identify them. Since the number of such judgements are relatively small, their effect can be mitigated after appropriate aggregation of multiple judgements. In addition, they could be a useful resource for studying whether the disagreements are caused by \textit{annotation error} or \textit{genuine difficulty} in the annotation, as suggested in \citet{poesio-etal-2019-crowdsourced}.

We also found that the exact match rate is highest when the number of referents is 1, and much lower as the number of referents increases. This is reasonable because referring expressions for multiple entities tend to be more pragmatic and ambiguous (e.g. \textit{``a cluster"}, \textit{``most of"}, \textit{``a line"}), and it would be more difficult to match the referents exactly. Note that entity level agreements are still at a high level, and the interpreted referents seem to mostly overlap with each other.

Next, as a preliminary analysis to study which expressions tend to have higher (or lower) disagreements, we compute the correlations between the occurrence of common tokens (represented by binary values) and the exact match rate of the pairwise judgements for each markable. Illustrative examples are shown in Table \ref{token_agreement_corrleation} and the whole list will be publicly available.

\begin{table}[htb!]
\centering \small
\setlength{\tabcolsep}{0.6em}
\begin{tabular}{cc}
\begin{tabular}{c|cc}
\toprule
Low & $\rho$ & Count \\
\midrule
\textit{it} & -0.149 & 12.7K \\
\textit{any} & -0.103 & \phantom{0}0.5K \\
\textit{that} & -0.100 & 12.5K \\
\textit{your} & -0.083 & \phantom{0}1.5K \\
\textit{few} & -0.081 & \phantom{0}0.1K \\
\textit{what} & -0.081 & \phantom{0}0.4K \\
\textit{others} & -0.064 & \phantom{0}0.8K \\
\textit{line} & -0.062 & \phantom{0}1.7K \\
\textit{bunch} & -0.060 & \phantom{0}0.2K \\
\textit{all} & -0.048 & \phantom{0}1.1K \\
\textit{triangle} & -0.046 & \phantom{0}2.5K \\
\textit{some} & -0.042 & \phantom{0}0.2K \\
\textit{medium} & -0.041 & 12.5K \\
\textit{another} & -0.039 & \phantom{0}1.4K \\
\textit{and} & -0.029 & \phantom{0}1.7K \\
\bottomrule
\end{tabular}
&
\begin{tabular}{c|cc}
\toprule
High & $\rho$ & Count \\
\midrule
\textit{lower} & 0.028 & \phantom{0}1.3K \\
\textit{two} & 0.030 & 14.7K \\
\textit{three} & 0.031 & \phantom{0}4.2K \\
\textit{darkest} & 0.036 & \phantom{0}2.1K \\
\textit{larger} & 0.039 & \phantom{0}7.7K \\
\textit{middle} & 0.041 & \phantom{0}2.1K \\
\textit{smallest} & 0.043 & \phantom{0}2.0K \\
\textit{very} & 0.056 & \phantom{0}6.1K \\
\textit{top} & 0.061 & \phantom{0}5.2K \\
\textit{light} & 0.072 & 18.7K \\
\textit{tiny} & 0.076 & \phantom{0}7.8K \\
\textit{large} & 0.084 & 21.7K \\
\textit{the} & 0.125 & 55.0K \\
\textit{one} & 0.136 & 57.1K \\
\textit{black} & 0.145 & 26.9K \\
\bottomrule
\end{tabular}
\end{tabular}
\caption{\label{token_agreement_corrleation}
Tokens with low or high correlation with the exact match rate. Correlation scores are shown in $\rho$.
}
\end{table}

In general, the correlations are very small and the amount of disagreements seem relatively constant across all token types. However, the general trend is still intuitive: ambiguous or complex expressions such as pronouns, interrogatives, quantifiers, and coordinating conjunctions tend to have negative correlations, while simple and less ambiguous expressions tend to have positive correlations.

To summarize the analyses, our annotation has high overall agreement but also includes interesting, reasonable disagreements which capture the ambiguity and uncertainty under continuous and partially-observable context.

\subsection{Pragmatic Expressions}
\label{subsection:pragmatic_expressions}

Finally, as an illustrative example of additional analyses that can be conducted based on our annotation, we give a more quantitative analysis of \textit{pragmatic expressions} which have been pointed out to exist in previous work but without sufficient amount of evidence \cite{udagawa2019natural}.

\begin{figure}[ht]
\centering
\includegraphics[width=0.45\textwidth]{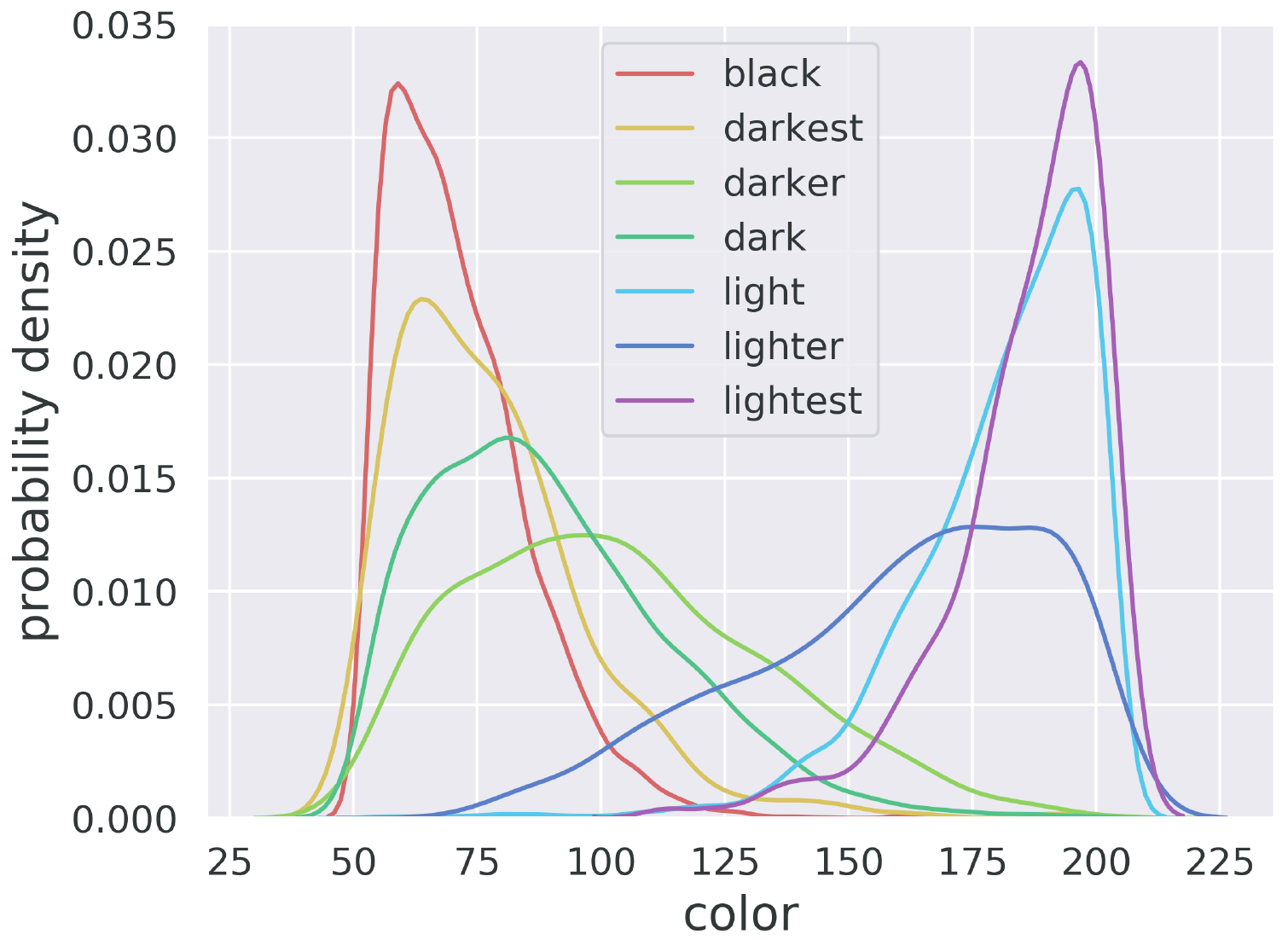}
\caption{Distribution of the actual color of the referents expressed by common adjectives (the range of color is 256 as in RGB scale, lower is darker).}
\label{fig:color_distplot}
\end{figure}

In this work, we focus on pragmatic expression of \textit{color} and estimate the distribution of the actual color of the referents described by the common adjectives. We simply assume that the adjective in the minimal noun phrase describe the color of the referents, since the exceptions (such as negation in the prenominal modifier) seemed rare and ignorable. Distributions are calculated based on kernel density estimation. As we can see in Figure \ref{fig:color_distplot}, all adjectives (including the specific color \textit{black}) have smooth and wide distributions which overlap with each other. This is a strong evidence that the same color can be described in various ways and become more pragmatic under continuous context.

\section{Experiments}
\label{section:experiments}

In this section, we evaluate and analyze baseline models based on three tasks. First is the \textit{target selection} task proposed by \citet{udagawa2019natural}, which tries to predict the entity selected by each player at the end of the collaborative referring task: this requires correct recognition of the created common ground based on the dialogue and context (i.e. player's view). Second is the \textit{reference resolution} task, where we focus on binary predictions of whether each entity is included in the referents or not. Last is the \textit{selfplay dialogue} task where the model plays the whole collaborative referring task (Section \ref{section:background_task}) against an identical copy of itself.

For reference resolution, we use simple majority voting (at the entity level) and automatic annotation of the referents to create gold annotation. Markables are removed if the majority considered them as unidentifiable.

\subsection{Model Architecture}
\label{subsection:model_architecture}

\begin{figure*}[tb!]
\centering
\includegraphics[width=0.90\textwidth]{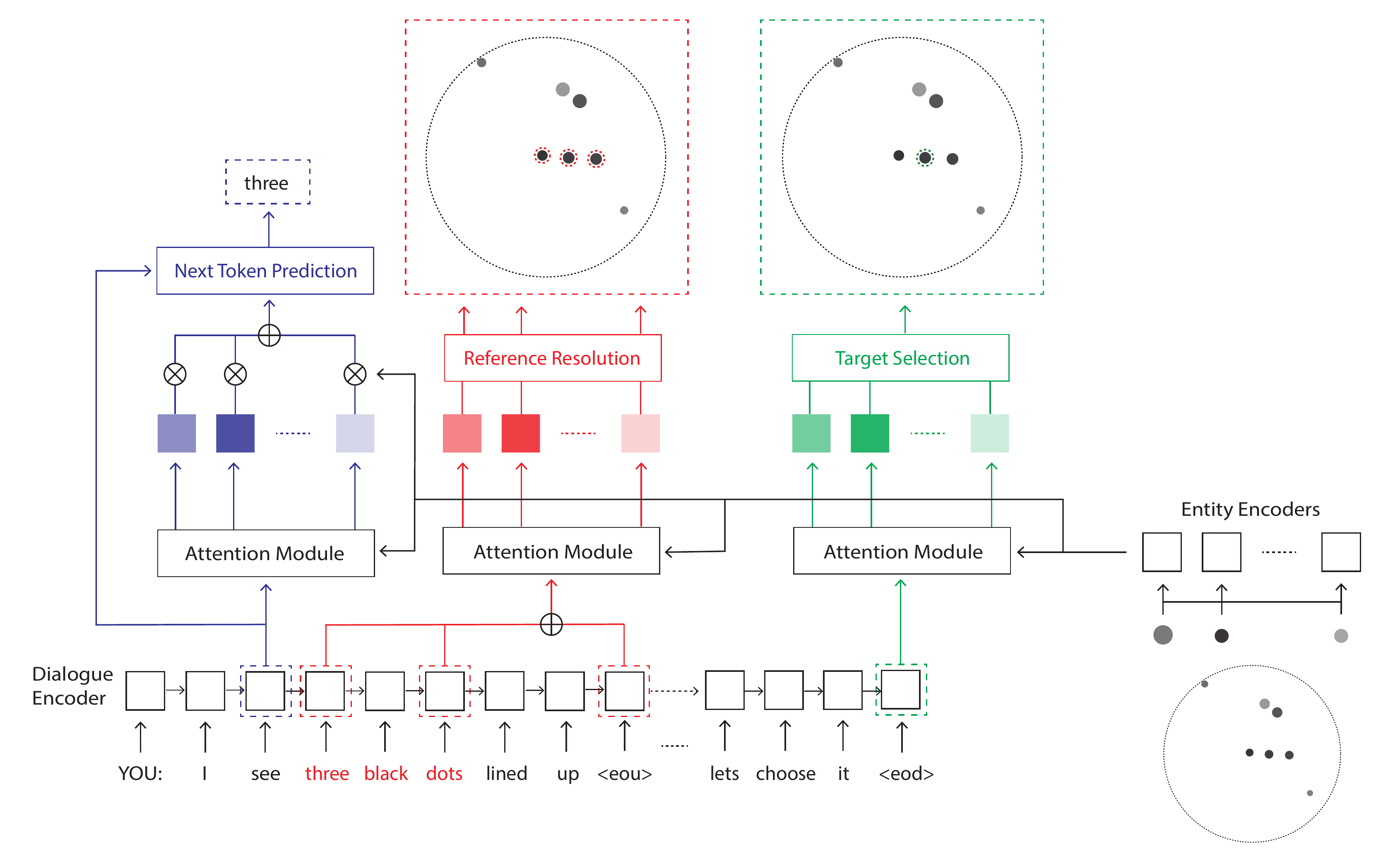}
\caption{Our baseline model architecture (best seen in color). \texttt{TSEL} decoder is shown in green, \texttt{REF} decoder and the input markable \textit{three black dots} are in red, and \texttt{DIAL} decoder is in blue. All decoders share the entity-level attention module.}
\label{fig:model_architecture}
\end{figure*}

The overall architecture of our baseline models is shown in Figure \ref{fig:model_architecture}.

\subsubsection{Encoders}

Our baseline models have two encoders: one for encoding dialogue tokens and one for context information.

Dialogue tokens are encoded with a standard GRU \cite{cho2014properties}. To encode context information, we embed each entity using a shared \textit{entity encoder}. This consists of an \textit{attribute encoder} which embeds the attributes of each entity (size, color and location) with a matrix followed by a tanh layer, and a \textit{relational encoder} which embeds relative attributes of each entity pairs (e.g. distance) with another matrix followed by a tanh layer. The final embedding of each entity is the concatenation of its attribute embedding and the sum of relational embeddings with the other 6 entities.

\subsubsection{Decoders}

Our models can have up to three decoders: \texttt{TSEL} for target selection, \texttt{REF} for reference resolution, and \texttt{DIAL} for predicting next tokens. Each decoder shares (some or all layers of) the \textit{attention module} based on MLP to compute a scalar score for each entity based on its embedding and certain positions of the GRU: \texttt{TSEL} takes the final hidden state, \texttt{REF} takes (the mean of) the start position of the markable, the end position of the markable, and the end position of the utterance including the markable, and \texttt{DIAL} takes the current hidden state. Based on these attention scores, \texttt{TSEL} simply computes the softmax and \texttt{REF} computes logistic regressions for each entity. \texttt{DIAL} reweights the entity embeddings based on these attention scores, concatenates it with the current hidden state and decodes with an MLP \cite{bahdanau2014nmt}.\\

In this experiment, we built five models based on different combinations of the three decoders. All models are trained with the default hyperparameters with minimal tuning.

\subsection{Results}
\label{subsection:results}

\begin{table*}[tb!]
\centering \small
\begin{tabular}{c|cccccc}
\toprule
\multirow{2}{*}{Model} & \multirow{2}{*}{Target Selection} & \multirow{2}{*}{Reference Resolution (Exact Match)} & \multicolumn{3}{c}{Selfplay Dialogue} \\
 & & & \#Shared=4 & \#Shared=5 & \#Shared=6 \\
\midrule
\texttt{TSEL} & 67.79$\pm$1.53  & -  & - & - & - \\
\texttt{REF} & - & \textbf{85.75$\pm$0.22} (\textbf{33.91$\pm$0.86}) & - & - & - \\
\texttt{TSEL-REF} & \textbf{69.01$\pm$1.58} & 85.47$\pm$0.36 (32.88$\pm$1.28) & - & - & - \\
\midrule
\texttt{TSEL-DIAL} & 67.01$\pm$1.29 & - & 42.07$\pm$1.27 & 57.37$\pm$1.29 & 77.00$\pm$1.13 \\
\texttt{TSEL-REF-DIAL} & \textbf{69.09$\pm$1.12} & \textbf{85.86$\pm$0.18} (\textbf{33.66$\pm$0.93}) & \textbf{45.78$\pm$2.15} & \textbf{61.95$\pm$1.72} & \textbf{80.01$\pm$1.61} \\
\midrule
Human & 90.79 & 96.26 (86.90) & 65.83 & 76.96 & 87.00 \\
\bottomrule
\end{tabular}
\caption{\label{baseline_results}
Results of our baseline models. Human scores from \citet{udagawa2019natural} and Table \ref{referent_identification_statistics} as a reference.
}
\end{table*}

We run the experiments 10 times with different random seeds and dataset splits (8:1:1 for train, validation and test). For selfplay dialogues, we generated 1,000 scenarios with each number of shared entities (4, 5 or 6) and set the output temperature to 0.25 during next token prediction. We report the mean and standard deviation of the results in Table \ref{baseline_results}.

In terms of \textit{target selection} and \textit{selfplay dialogue} tasks, we found consistent improvements by training the models jointly with reference resolution. This verified that we can indeed leverage the central subtask of reference resolution to improve performance on difficult end tasks. The results for \textit{reference resolution} are reasonably high in terms of entity level accuracy but much lower in terms of exact match rate. Considering the high agreements (Subsection \ref{subsection:agreement_statistics}) and improved reliability of the gold annotation after aggregation, we expect there to be a huge room for further improvements.

Overall, common grounding under continuous and partially-observable context is still a challenging task, and we expect our resource to be a fundamental testbed for solving this task through advanced skills of reference resolution.

\subsection{Analysis}
\label{subsection:analysis}

To demonstrate the advantages of our approach for interpreting and analyzing dialogue systems, we give a more detailed analysis of \texttt{TSEL-REF-DIAL} model which performed well on all three tasks. In Table \ref{exact_match_results}, we show the results for reference resolution (entity level accuracy and exact match rate) grouped by the number of referents in the gold annotation. In terms of the exact match rate, we found that the model performs very well on 0 and 7 referents: this is because most of them can be recognized at the superficial level, such as \textit{``\underline{none of them}"}, \textit{``\underline{all of mine}"} or \textit{``I don't have \underline{that}"}. However, the model struggles on all other cases: the results are especially worse for markables with more than 1 referent. This shows that the model still lacks the ability of precisely tracking multiple referents, which can be expressed in complex, pragmatic ways (such as groupings).

\begin{table}[tb!]
\centering \small
\begin{tabular}{c|ccc}
\toprule
\# Referents & \% Accuracy & \% Exact Match & Count \\
\midrule
0 & 95.91$\pm$1.38 & 83.53$\pm$4.65\phantom{0} & \phantom{0}148.5 \\
1 & 89.34$\pm$0.17 & 36.86$\pm$1.32\phantom{0} & 2782.5 \\
2 & 78.14$\pm$1.07 & 20.59$\pm$1.90\phantom{0} & \phantom{0}587.9 \\
3 & 70.64$\pm$1.02 & 13.63$\pm$2.06\phantom{0} & \phantom{0}283.3 \\
4 & 69.12$\pm$2.69 & 10.16$\pm$3.47\phantom{0} & \phantom{00}81.0 \\
5 & 73.57$\pm$2.94 & 17.56$\pm$5.88\phantom{0} & \phantom{00}33.0 \\
6 & 78.69$\pm$4.45 & 13.18$\pm$7.31\phantom{0} & \phantom{00}43.0 \\
7 & 74.60$\pm$7.49 & 50.38$\pm$11.40 & \phantom{00}22.3 \\
\bottomrule
\end{tabular}
\caption{\label{exact_match_results}
Results of the reference resolution task grouped by the number of referents in the gold annotation (along with the average count of such markables in the test set).
}
\end{table}

In addition, we found that the correlation between reference resolution score (average accuracy of reference resolution in each dialogue) and target selection score (binary result of target selection in each dialogue) was relatively weak, with an average of only $0.23$ in 10 runs of the experiments. Indeed, we verified that the model is often correct for the target selection task based on the \textit{wrong reason}, without tracking the referents correctly. Our annotation is also useful for \textit{error analysis} in recognizing the process of common grounding, by inspecting where the model made a mistake and lost track of the correct referents.


\begin{figure}[tb!]
\centering
\begin{tikzpicture}
\node[inner sep=0pt] (agent_0) at (0,0)
  {\includegraphics[width=0.23\textwidth]{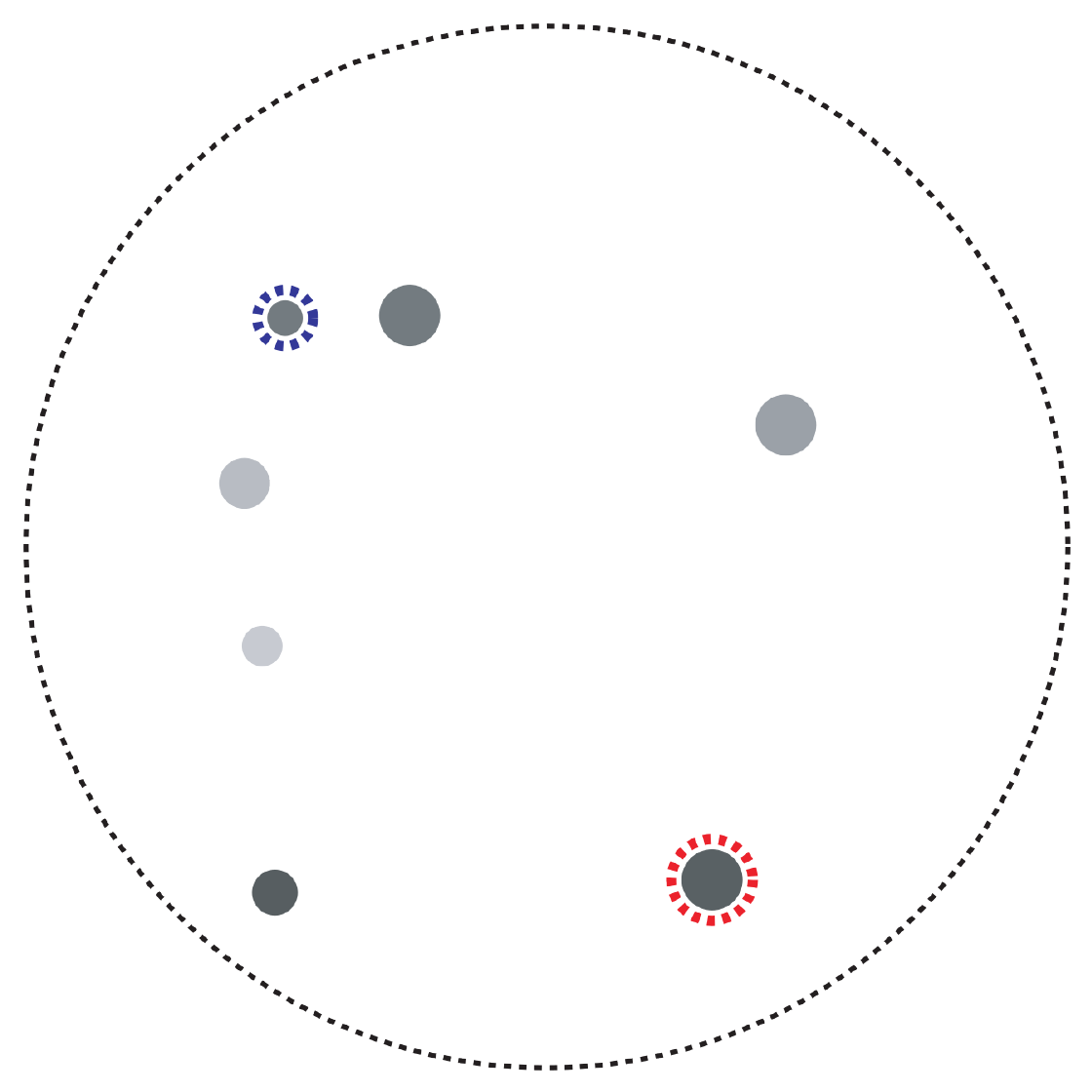}};
\node[inner sep=0pt] (agent_1) at (4.2,0)
  {\includegraphics[width=0.23\textwidth]{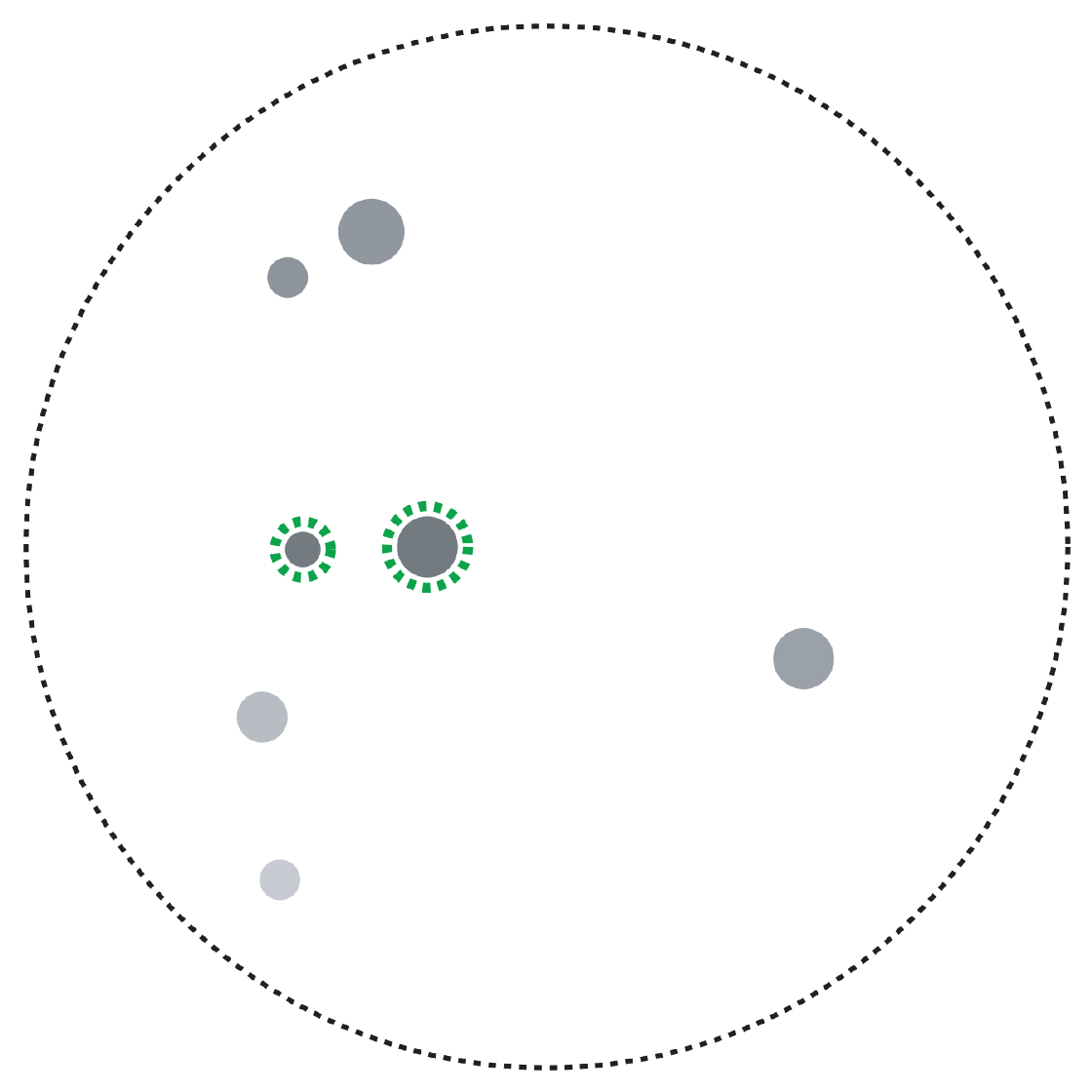}};
\node [below] at (0,-2) {Model A's view};
\node [below] at (4.2,-2) {Model B's view};
\end{tikzpicture}
\small
\begin{tabular}{l}
\toprule
A: I have {\color{myred} \underline{\textbf{a large black dot}}} with {\color{myblue} \underline{\textbf{a smaller dark dot}}} to the\\ right of {\color{myred} \underline{\textbf{it}}} \\
B: I see {\color{mygreen} \underline{\textbf{that}}} . Let's pick \underline{the large black dot} \\
\bottomrule
\end{tabular}
\caption{Example dialogue from the selfplay task by \texttt{TSEL-REF-DIAL} model. Predicted referents are highlighted (no referents were predicted for \textit{the large black dot}).
}
\label{fig:selfplay}
\end{figure}

Finally, we show an example dialogue from the selfplay task along with the interpreted process of common grounding in Figure \ref{fig:selfplay}. Referring expressions are automatically detected by a BiLSTM-CRF tagger \cite{huang2015bidirectional} trained on our corpus (with 98.9\% accuracy at the token level). Based on the raw dialogue only, it is difficult to identify which dots the models are referring to. However, by visualizing the intended referents, we can see that model A is describing two dots in somewhat unnatural and inappropriate way (albeit using the anaphoric expression \textit{it} appropriately). In turn, model B acknowledges this in a perfectly coherent way but without predicting any referents for \textit{the large black dot}: we often observed such phenomena, where the utterance by a model cannot be interpreted correctly \textit{even by itself}. This way, our annotation allows for fine-grained analysis of both \textit{capabilities} and \textit{incapabilities} of existing dialogue systems. The generated dialogue is short in this example, but our approach would be even more critical for interpretation as the dialogues get longer and more complicated.

\section{Conclusion}
\label{section:conclusion}

We propose a novel method of decomposing common grounding based on its subtasks to study the intermediate process of common grounding. We demonstrated the advantages of our approach through extensive analysis of the annotated corpus and the baseline models. Overall, we expect our work to be a fundamental step towards interpreting and improving common grounding through reference resolution.

\section*{Acknowledgements}
This work was supported by JSPS KAKENHI Grant Number 18H03297 and NEDO SIP-2 ``Big-data and AI-enabled Cyberspace Technologies.” We also thank the anonymous reviewers for their valuable suggestions and comments.

\fontsize{9.0pt}{10.0pt} \selectfont

\bibliographystyle{aaai}
\bibliography{3644.myref}

\end{document}